\documentclass{article}

\usepackage{arxiv}
\usepackage[utf8]{inputenc}
\usepackage[T1]{fontenc}
\usepackage{hyperref}
\usepackage{url}
\usepackage{booktabs}
\usepackage{amsfonts}
\usepackage{amsmath}
\usepackage{nicefrac}
\usepackage{microtype}
\usepackage{graphicx}
\usepackage{array}
\graphicspath{ {./images/} }

\hypersetup{
  colorlinks=true,
  linkcolor=blue,
  citecolor=blue,
  urlcolor=blue
}

\title{Literature-Guided Minimax Optimization of Virtual Epilepsy Neurostimulation}

\author{
  Cathy Liu
}

\begin{document}
\maketitle

\begin{abstract}
Computational models of epilepsy promise patient-specific treatment design, but most optimization workflows still search for parameters that perform well on average. In neuromodulation, this is a weak target: a protocol that improves the mean response can still fail in the patient whose network is least tolerant to stimulation. We present a literature-guided minimax pipeline that couples PubMed-scale hypothesis extraction, The Virtual Brain (TVB) Epileptor simulations, and large-language-model-guided black-box optimization. The optimizer proposes either intrinsic model-control parameters or clinically interpretable external-stimulation protocols; TVB evaluates each proposal across sampled virtual patients; and the objective maximizes worst-case reward, defined as the negative variance of simulated seizure activity. In the intrinsic model-control positive-control experiment, the best archived parameter set improved worst-case reward from $-0.5285$ to $-0.3182$, a $39.8\%$ gain over baseline, and maintained large gains in held-out and stress-test virtual cohorts. The clinical-style external-stimulation search produced a smaller archived improvement, but a new all-region stimulation landscape over 760 candidates showed that the selected right-hippocampal protocol ranked fourth overall and was within $0.0082$ reward units of the grid optimum. We use this intentionally small grid-searchable space as a calibration setting: it tests whether literature-guided proposals land near high-performing regions before moving to open-ended stimulation spaces where exhaustive search over targets, waveforms, timing, and patient states is infeasible. A 20-patient virtual cohort still showed no aggregate benefit of the right-hippocampal protocol ($p=0.9019$), despite a 55\% responder rate. These results support LLMs as sample-efficient hypothesis generators inside a simulation loop, not as clinical decision-makers or proven optimizer replacements.
\end{abstract}

\section{Introduction}
Drug-resistant epilepsy is a systems problem as much as a focal pathology. Seizures arise from nonlinear interactions among local excitability, long-range structural connectivity, and patient-specific seizure onset networks. Responsive neurostimulation and deep brain stimulation have established that electrical intervention can reduce seizure burden in selected patients, but their efficacy remains heterogeneous: electrode location, stimulation amplitude, disease subtype, and latent brain state all modulate response. A computational optimizer for epilepsy treatment therefore faces a demanding target. It must not merely improve a representative patient; it must avoid catastrophic failure in the patient whose network lies at the edge of the sampled distribution.

Virtual brain models provide a principled way to study this problem before clinical deployment. The Virtual Brain (TVB) platform simulates whole-brain neural mass dynamics constrained by empirical connectomes, while the Epileptor model captures seizure initiation, propagation, and termination using a compact dynamical-system representation \cite{sanzleon2013tvb,jirsa2014seizure,proix2017vep}. These tools make it possible to evaluate candidate interventions across virtual cohorts, but they do not by themselves solve the search problem. A realistic intervention space contains continuous parameters, discrete targets, mechanistic priors from the literature, and strong safety constraints. Exhaustive search is expensive; naive average-case optimization is clinically fragile.

This project asks whether a large language model (LLM) can act as a literature-aware proposal engine inside a rigorous simulation loop. The LLM does not evaluate outcomes and is not treated as a scientific oracle. Instead, it proposes candidate parameters, grounded in a corpus of epilepsy and neuromodulation abstracts. TVB supplies the reward. The optimization criterion is minimax robustness: choose the protocol whose worst virtual-patient outcome is best.

The methodological motivation is sample efficiency under scientific structure. In this paper's constrained external-stimulation experiment, the search space is deliberately small enough to evaluate exhaustively: 76 regions times 10 stimulation intensities. That makes it a useful calibration problem because the LLM-selected candidate can be ranked against the true evaluated landscape. Real neurostimulation design is not this small. Once waveform shape, frequency, pulse width, phase, electrode montage, closed-loop timing, disease subtype, and patient state are included, exhaustive grid search becomes combinatorially impractical. The intended role for language-model proposals is therefore not to replace exhaustive search in small spaces, but to provide literature-informed candidates when the experimentally meaningful space is too large or too semantic for naive enumeration.

The central result is intentionally bifurcated. When the optimizer is allowed to tune intrinsic model parameters controlling epileptogenicity and coupling, it finds a strong robust solution: worst-case reward improves by 39.8\% in the archived optimization trajectory. When the same framework is restricted to a more clinically interpretable external-current boost at one brain region, improvement is much smaller and cohort-level effects are mixed. That contrast is scientifically useful. It suggests that the pipeline can discover robust dynamical regimes, while also revealing how much harder it is to translate those regimes into a constrained stimulation protocol.

The contributions of this work are fourfold:
\begin{enumerate}
  \item a reproducible research-to-simulation-to-optimization pipeline linking PubMed-scale literature mining, TVB Epileptor simulation, and LLM-guided minimax search;
  \item a robust intrinsic-control result showing large worst-case improvement across virtual patients;
  \item a translational stress test and all-region stimulation landscape showing that external-stimulation effects are heterogeneous, optimizer claims require calibration, and patient-response variability is substantial; and
  \item an open static research essay, codebase, and paper artifact designed for independent inspection and extension.
\end{enumerate}

\subsection{Scope of claims}
The claims are intentionally bounded. This project demonstrates an auditable in silico workflow for converting literature-derived hypotheses into minimax TVB experiments. It does not demonstrate clinical efficacy, validate a stimulation protocol, or prove that LLM-guided optimization dominates Bayesian optimization or random search in general. The strongest optimizer claim is narrower: in a bounded anatomical search space where exhaustive calibration is possible, the literature-guided rHC candidate lands near the top of the evaluated landscape and substantially above the median random eight-evaluation search. That is evidence for sample-efficient hypothesis generation, not evidence for autonomous clinical optimization.

\section{Background}
\subsection{Virtual epilepsy patients}
Patient-specific computational epilepsy modeling has advanced from explanatory dynamical systems toward intervention planning. The Virtual Epileptic Patient framework combines structural connectivity, hypothesized epileptogenic zones, and neural mass models to simulate seizure spread and candidate interventions \cite{proix2017vep}. At the node level, the Epileptor model formalizes seizure onset and offset as transitions in a multi-timescale dynamical system \cite{jirsa2014seizure}. These models are especially attractive for neurostimulation because they allow parameter perturbations to be tested repeatedly under controlled assumptions.

The major limitation is that the virtual patient is not a real patient. Simulation results inherit assumptions about connectivity, node dynamics, observation windows, and parameterization. This work therefore treats TVB as a hypothesis engine: a platform for asking whether an optimization idea is coherent and stress-testable, not a substitute for electrophysiology, imaging, or clinical trials.

\subsection{Robust optimization for neuromodulation}
Clinical neuromodulation is a worst-case-sensitive domain. A parameter set that helps many patients but worsens a minority may be unacceptable, especially when adverse responses are difficult to predict. Minimax optimization makes this concern explicit. Instead of optimizing the mean response, it maximizes the minimum reward across a sampled cohort. The cost is conservatism; the benefit is that failures are directly visible in the objective.

For expensive black-box simulations, Bayesian optimization is a common baseline because it uses a surrogate model to reduce the number of evaluations \cite{snoek2012bayesopt}. The premise here is complementary. Literature-aware LLM proposals may be useful when the search space contains named brain regions, mechanistic priors, and qualitative constraints that are awkward to encode in a standard kernel. The LLM is not trusted to be correct; it is used to generate informed guesses that the simulator can accept or reject.

\section{Study Design}
\begin{figure}[t]
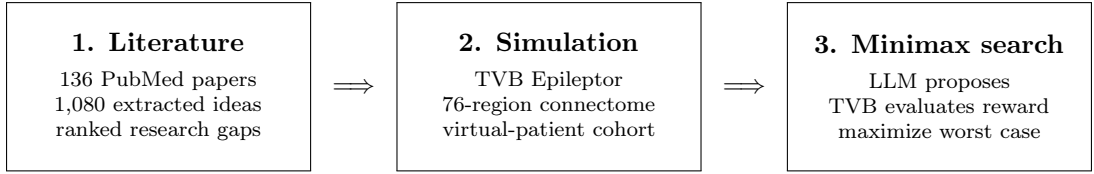

  \centering
  \setlength{\tabcolsep}{4pt}
  \newcommand{\pipelinebox}[2]{%
    \fbox{\parbox[c][2.0cm][c]{0.23\textwidth}{%
      \centering \textbf{#1}\\[0.35em]\footnotesize #2%
    }}%
  }
  \begin{tabular}{ccccc}
    \pipelinebox{1. Literature}{136 PubMed papers\\1,080 extracted ideas\\ranked research gaps}
    &
    $\Longrightarrow$
    &
    \pipelinebox{2. Simulation}{TVB Epileptor\\76-region connectome\\virtual-patient cohort}
    &
    $\Longrightarrow$
    &
    \pipelinebox{3. Minimax search}{LLM proposes\\TVB evaluates reward\\maximize worst case}
  \end{tabular}
  \caption{\textbf{Pipeline schematic.} A PubMed corpus is converted into structured research gaps, the highest-priority robust-transfer question is instantiated in a TVB Epileptor virtual cohort, and an LLM proposes candidate parameters that are accepted only after simulation-based reward evaluation. The objective is minimax: improve the least favorable sampled virtual-patient response.}
  \label{fig:method}
\end{figure}

Figure~\ref{fig:method} summarizes the pipeline. The study begins by mining epilepsy, digital-twin, and neurostimulation literature from PubMed. The extracted research gaps motivate a robust-transfer problem: can an intervention selected in simulation improve the worst virtual-patient outcome rather than merely the average outcome? Candidate interventions are then proposed by an LLM and evaluated by TVB. All reported numerical results in this manuscript are taken from checked-in repository artifacts; no new scientific result is introduced by the paper-generation script.

\section{Methods}
\subsection{Literature mining and hypothesis selection}
The literature stage queried PubMed for work at the intersection of epilepsy, whole-brain modeling, stimulation, digital twins, and responsive neurostimulation. Deduplication was performed by exact PMID, yielding 136 unique papers. Titles and abstracts were passed to an LLM extraction prompt that returned structured open questions, future experiments, limitations, and TVB relevance. The final artifact contains 1,080 extracted ideas. A second LLM stage grouped ideas into thematic clusters and scored each cluster on novelty, TVB feasibility, and clinical impact. The robust-transfer study was selected because it scored highly on all three axes and could be tested directly in TVB.

This stage is a hypothesis-generation procedure. No claim is made that the extracted gaps are complete, deterministic, or expert validated. The role of literature mining is to bias search toward plausible questions; the role of simulation is to impose quantitative discipline.

\subsection{Virtual brain model}
Simulations use the TVB Epileptor model on the 76-region connectivity distributed with the TVB library. The repository loads this atlas with \texttt{Connectivity.from\_file()}, so region indices follow TVB's built-in order rather than a custom parcellation. Table~\ref{tab:regionmap} reports the index-to-label mapping used by the optimization prompts and result artifacts.

For intrinsic-control experiments, candidate parameters are epileptogenicity $x_0$ and linear coupling $K$. The baseline configuration is $x_0=-1.6$ and $K=0.0152$. The optimized intrinsic setting found in the archived trajectory is $x_0=-2.1$ and $K=0.0165$. Apart from the candidate values of $x_0$ and the scalar linear coupling, the intrinsic simulations use TVB's default \texttt{Epileptor()} parameterization.

For clinical-style external-stimulation experiments, the model holds intrinsic parameters fixed and applies an external-current boost $b$ to one selected region. The candidate protocol is therefore $\theta=(b,s)$, where $b\in[0,4]$ and $s\in\{0,\ldots,75\}$ is a brain-region index. The best archived clinical-style recommendation is a low-intensity boost $b=0.6$ at the right hippocampus (rHC, index 9).

\begin{table}[t]
  \centering
  \scriptsize
  \caption{\textbf{TVB 76-region index map.} Region labels are loaded from the default TVB connectivity object. Prefixes \texttt{r} and \texttt{l} denote right and left hemisphere labels.}
  \label{tab:regionmap}
  \begin{tabular}{r@{\ }l r@{\ }l r@{\ }l r@{\ }l}
    \toprule
    \multicolumn{2}{c}{Index} & \multicolumn{2}{c}{Index} & \multicolumn{2}{c}{Index} & \multicolumn{2}{c}{Index} \\
    \midrule
    0 & rA1 & 19 & rPFCDM & 38 & lA1 & 57 & lPFCDM \\
    1 & rA2 & 20 & rPFCM & 39 & lA2 & 58 & lPFCM \\
    2 & rAMYG & 21 & rPFCORB & 40 & lAMYG & 59 & lPFCORB \\
    3 & rCCA & 22 & rPFCPOL & 41 & lCCA & 60 & lPFCPOL \\
    4 & rCCP & 23 & rPFCVL & 42 & lCCP & 61 & lPFCVL \\
    5 & rCCR & 24 & rPHC & 43 & lCCR & 62 & lPHC \\
    6 & rCCS & 25 & rPMCDL & 44 & lCCS & 63 & lPMCDL \\
    7 & rFEF & 26 & rPMCM & 45 & lFEF & 64 & lPMCM \\
    8 & rG & 27 & rPMCVL & 46 & lG & 65 & lPMCVL \\
    9 & rHC & 28 & rS1 & 47 & lHC & 66 & lS1 \\
    10 & rIA & 29 & rS2 & 48 & lIA & 67 & lS2 \\
    11 & rIP & 30 & rTCC & 49 & lIP & 68 & lTCC \\
    12 & rM1 & 31 & rTCI & 50 & lM1 & 69 & lTCI \\
    13 & rPCI & 32 & rTCPOL & 51 & lPCI & 70 & lTCPOL \\
    14 & rPCIP & 33 & rTCS & 52 & lPCIP & 71 & lTCS \\
    15 & rPCM & 34 & rTCV & 53 & lPCM & 72 & lTCV \\
    16 & rPCS & 35 & rV1 & 54 & lPCS & 73 & lV1 \\
    17 & rPFCCL & 36 & rV2 & 55 & lPFCCL & 74 & lV2 \\
    18 & rPFCDL & 37 & rCC & 56 & lPFCDL & 75 & lCC \\
    \bottomrule
  \end{tabular}
\end{table}

\subsection{Simulation implementation details}
Each simulation uses the deterministic Euler integrator with \texttt{dt=0.05}, a \texttt{TemporalAverage} monitor with period $1.0$, and \texttt{simulation\_length=1000.0} in the packaged simulation configuration. The current code does not explicitly discard a transient interval before computing reward; all returned temporal-average samples are included in the variance calculation. The simulator is deterministic: there is no stochastic integrator and no added process noise. Randomness enters only through virtual-patient generation.

For the five-patient minimax optimization cohort, virtual patients are generated by drawing a scalar coupling perturbation from $\mathcal{U}(-0.003,0.003)$ with seed 42 and adding it to the candidate coupling value. Thus, for an intrinsic candidate with coupling $K$, patient $p$ is evaluated at $K+\epsilon_p$. The same perturbation scheme is used for clinical-style robust evaluation, except the base coupling is fixed at $0.0152$. For the 20-patient cohort artifact, the code instead perturbs the full connectivity matrix by multiplicative normal noise with mean 1.0 and standard deviation 0.1, clips negative weights to zero, samples an SOZ class from the configured probabilities, and raises the SOZ node's $x_0$ by a uniform offset in $[0.1,0.3]$.

External stimulation is implemented as a direct change to the Epileptor external-current vector. The code initializes \texttt{Iext} to $3.1$ for all 76 regions and then sets \texttt{Iext[s] = 3.1 + b} for the selected site $s$. The rHC recommendation is therefore not a waveform or electrode model; it is an additive current offset at TVB label rHC, whose atlas index is 9. The bound $b\in[0,4]$ is enforced in the optimizer prompt and typed configuration as an operational search box. It should not be interpreted as a device-calibrated safety range. The repository labels $b<1.0$ as low intensity, $1.0\leq b<2.5$ as moderate intensity, and larger values as high intensity for reporting only.

\subsection{Reward function}
Let $x_{1,v}(t;\theta,p)$ denote the simulated fast Epileptor state at region $v$ and time $t$ for intervention parameters $\theta$ and virtual patient $p$. The seizure-burden score used by the code is the mean temporal variance across regions:
\begin{equation}
  S(\theta;p)=\frac{1}{|V|}\sum_{v\in V}\operatorname{Var}_{t}\left[x_{1,v}(t;\theta,p)\right].
\end{equation}
The reward is its negative,
\begin{equation}
  R(\theta;p)=-S(\theta;p),
\end{equation}
so higher reward corresponds to lower simulated seizure-intensity variance. This is a model endpoint, not a clinical seizure-frequency endpoint.

\subsection{Minimax objective}
For a sampled cohort $\mathcal{P}$ of virtual patients, the robust objective is
\begin{equation}
  \theta^{\star}=\arg\max_{\theta\in\Theta}\min_{p\in\mathcal{P}}R(\theta;p).
\end{equation}
The training optimization used five virtual patients generated by uniform coupling perturbations with random seed 42. Additional archived artifacts evaluate the selected intrinsic parameters on held-out patients, wider perturbation regimes, and a 30-patient stress distribution. The codebase records these as separate artifacts, but it does not contain provenance metadata proving that the held-out cohorts were locked before the optimization run. They should therefore be read as archived secondary validation artifacts rather than as a formally pre-registered test set.

\subsection{LLM-guided search}
The LLM search loop is deliberately simple. At each iteration, the model receives the prior optimization history and is asked to return a JSON object containing the next candidate parameters. TVB evaluates the candidate and appends the observed worst-case and mean rewards to the history. The intrinsic-control search tunes $(x_0,K)$; the clinical-style search tunes $(b,s)$. Both archived LLM trajectories contain eight evaluations. The first intrinsic evaluation is the baseline $(x_0=-1.6,K=0.0152)$ followed by seven LLM proposals; the clinical-style artifact contains eight external-stimulation evaluations. Optimization stops after this fixed evaluation budget. There is no convergence test, early stopping rule, or statistical stopping criterion. In the comparison artifact, a Gaussian-process Bayesian optimizer is run for the same number of external-stimulation evaluations and compared against the LLM-guided trajectory.

\subsection{External-stimulation landscape}
To calibrate the constrained-stimulation result against the full bounded search space, we added an all-region landscape experiment under the current \texttt{uv} environment with \texttt{tvb-library==2.10.0} and \texttt{tvb-data==3.0.0}. The experiment evaluates every TVB region at $b\in\{0.0,0.3,0.6,0.9,1.2,1.5,2.0,2.5,3.0,4.0\}$, yielding 760 stimulation candidates. Each candidate is evaluated on the same five-patient minimax cohort used by the clinical-style search. We then sample 100 random-search runs of budget eight from the evaluated grid, so random baselines are compared against identical simulator outputs rather than rerunning noisy simulations.

This landscape is a calibration artifact rather than a replacement for the original LLM optimization trace. The absolute rewards differ from the older checked-in clinical trace because the current reproducible environment explicitly installs TVB's data package and uses the currently resolved TVB runtime. For that reason, the landscape analysis emphasizes rank, gap to grid optimum, and relative comparison among candidates evaluated in the same environment. The point of using a grid-searchable space is methodological: it lets us test whether a literature-guided proposal is plausible when the answer can still be audited. In larger stimulation-design spaces, this audit would usually be unavailable.

\subsection{Virtual cohort analysis}
The 20-patient cohort artifact compares no stimulation with the selected clinical-style rHC protocol ($b=0.6$). Virtual seizure onset zone (SOZ) classes are sampled among hippocampal, frontal, temporal, and occipital categories. The analysis reports paired rewards, a paired $t$ statistic, a two-sided $p$ value, a standardized effect size, and responder/paradoxical-response rates. Because the cohort is simulated and small, these statistics are descriptive rather than inferential evidence for clinical use.

\section{Results}
\subsection{Intrinsic minimax control produced a large robust improvement}
\begin{figure}[t]
  \centering
  \includegraphics[width=\linewidth]{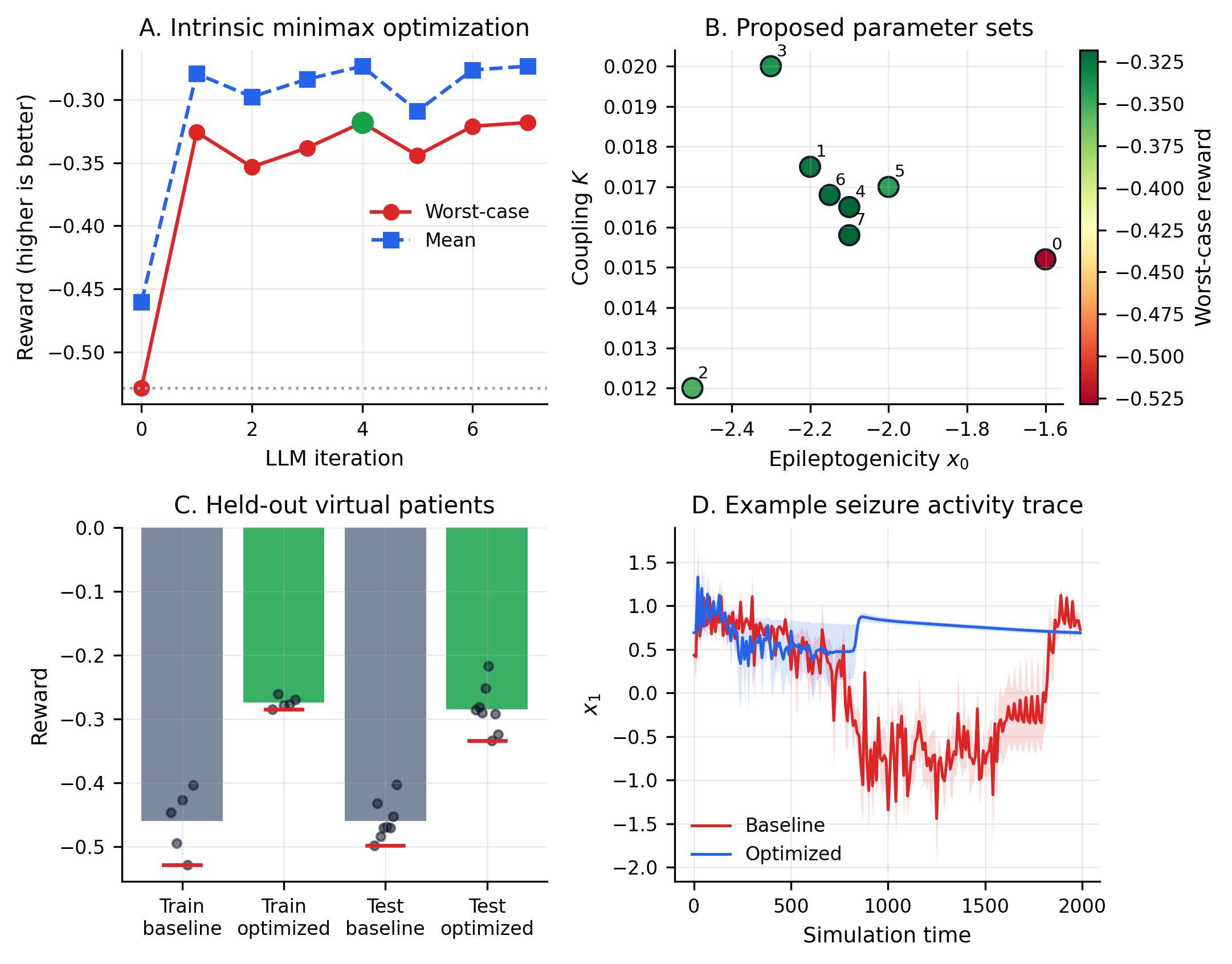}
  \caption{\textbf{Intrinsic minimax optimization.} \textbf{A}, LLM-guided optimization of epileptogenicity and coupling improved worst-case and mean reward relative to baseline. \textbf{B}, Proposed parameter sets concentrate around a lower-$x_0$ regime with slightly adjusted coupling. \textbf{C}, The optimized setting improves reward distributions in both training and held-out virtual patients; red horizontal marks denote worst-case rewards. \textbf{D}, Example simulated activity traces show lower mean fast-state variance under the optimized intrinsic regime.}
  \label{fig:intrinsic}
\end{figure}

The intrinsic optimization rapidly moved away from the baseline epileptogenicity and coupling regime. The best archived iteration occurred at $x_0=-2.1$ and $K=0.0165$, improving worst-case reward from $-0.5285$ to $-0.3182$ (39.8\%) and mean reward from $-0.4604$ to $-0.2736$ (Figure~\ref{fig:intrinsic}A). The search trajectory is notable because several later proposals remained close to the same region of parameter space, suggesting that the simulator repeatedly rewarded a lower-epileptogenicity, slightly higher-coupling regime rather than a one-off artifact.

Held-out evaluation preserved the effect direction. In the archived test cohort, worst-case reward improved from $-0.4985$ to $-0.3346$, while mean reward improved from $-0.4602$ to $-0.2848$ (Table~\ref{tab:mainresults}). The 30-patient stress distribution showed the same qualitative pattern: optimized worst-case reward was $-0.4047$ versus $-0.5175$ for baseline, and optimized mean reward was $-0.3055$ versus $-0.4451$.

\begin{table}[t]
  \centering
  \caption{\textbf{Archived quantitative results.} Rewards are negative seizure-burden scores, so less negative values are better. Percent change is computed on worst-case reward unless otherwise stated.}
  \label{tab:mainresults}
  \begin{tabular}{llrrrr}
    \toprule
    Experiment & Condition & $n$ & Worst reward & Mean reward & Worst-case gain \\
    \midrule
    Intrinsic trace & Baseline & 5 & $-0.5285$ & $-0.4604$ & -- \\
    Intrinsic trace & Best LLM proposal & 5 & $-0.3182$ & $-0.2736$ & $39.8\%$ \\
    Held-out intrinsic & Baseline & 8 & $-0.4985$ & $-0.4602$ & -- \\
    Held-out intrinsic & Optimized & 8 & $-0.3346$ & $-0.2848$ & $32.9\%$ \\
    Stress intrinsic & Baseline & 30 & $-0.5175$ & $-0.4451$ & -- \\
    Stress intrinsic & Optimized & 30 & $-0.4047$ & $-0.3055$ & $21.8\%$ \\
    External stimulation & No stimulation & 5 & $-0.5285$ & -- & -- \\
    External stimulation & rHC, $b=0.6$ & 5 & $-0.5194$ & -- & $1.7\%$ \\
    \bottomrule
  \end{tabular}
\end{table}

\subsection{Constrained external stimulation produced modest gains}
\begin{figure}[t]
  \centering
  \includegraphics[width=\linewidth]{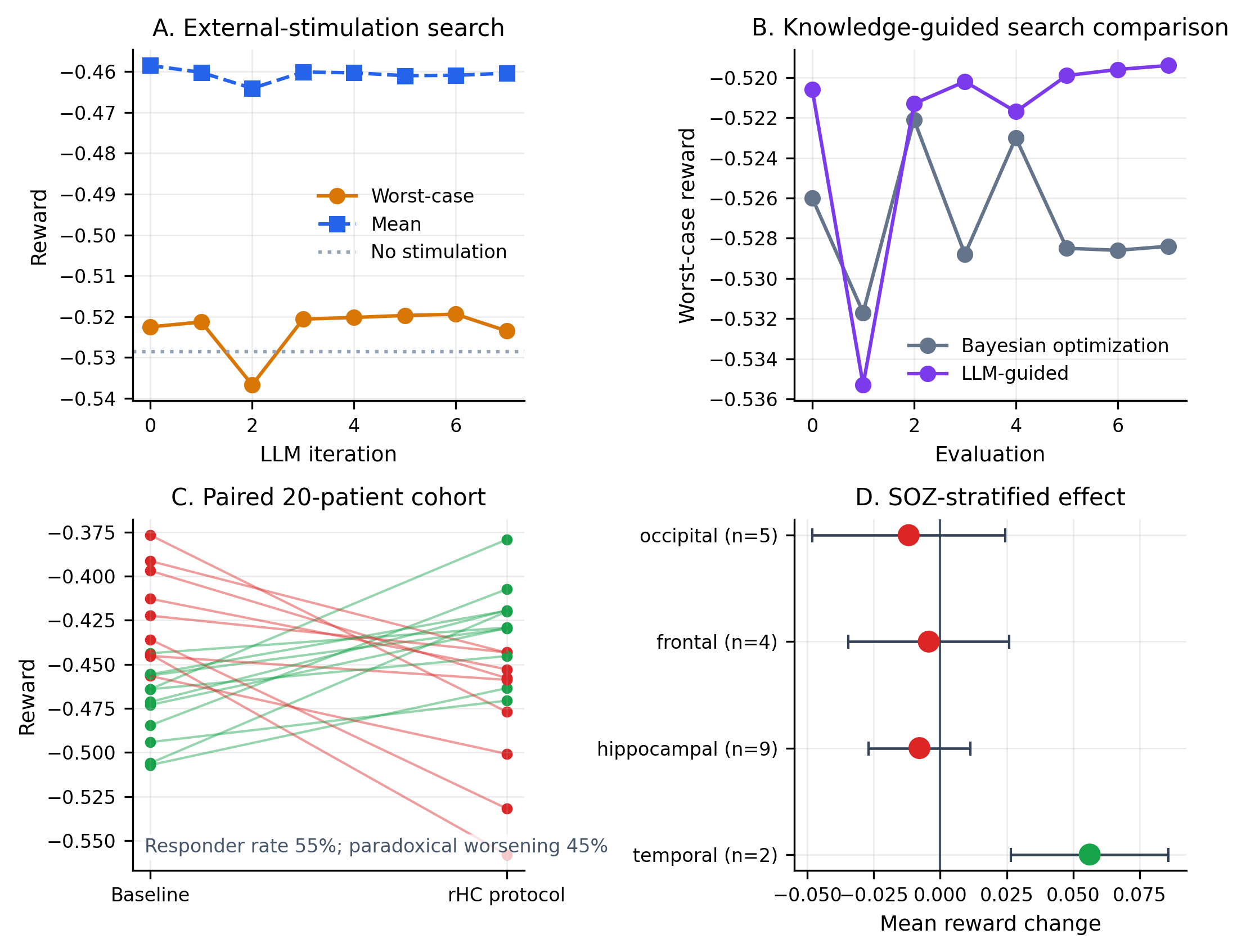}
  \caption{\textbf{Clinical-style stimulation and cohort response.} \textbf{A}, External-current stimulation produced only modest worst-case improvement relative to no stimulation. \textbf{B}, The LLM-guided trajectory slightly outperformed the Bayesian-optimization trajectory in the archived eight-evaluation comparison. \textbf{C}, In a 20-patient virtual cohort, the selected rHC protocol produced heterogeneous paired responses rather than a clear aggregate shift. \textbf{D}, SOZ-stratified analysis suggests a positive temporal subgroup signal but no consistent benefit across hippocampal, frontal, or occipital groups.}
  \label{fig:clinical}
\end{figure}

The external-stimulation search was harder. The best archived protocol was low-intensity rHC stimulation ($b=0.6$), improving worst-case reward from $-0.5285$ to $-0.5194$, or 1.7\% (Figure~\ref{fig:clinical}A). This is directionally positive but much smaller than the intrinsic-control result. The difference is expected: intrinsic control directly changes the dynamical parameters that govern seizure propensity, whereas external stimulation perturbs one region in an already coupled system.

The comparison with Bayesian optimization is informative but not definitive. Over eight evaluations, the LLM-guided external-stimulation trajectory reached a best reward of $-0.5194$, whereas the Bayesian optimizer reached $-0.5221$ (Figure~\ref{fig:clinical}B). The absolute margin is small ($0.0027$ reward units), but the LLM trajectory selected the literature-plausible rHC target immediately and improved from there. This supports the narrower claim that language-model priors can reduce wasted early evaluations in a structured, named-parameter search space.

\subsection{The all-region landscape calibrates the constrained search}
\begin{figure}[t]
  \centering
  \includegraphics[width=\linewidth]{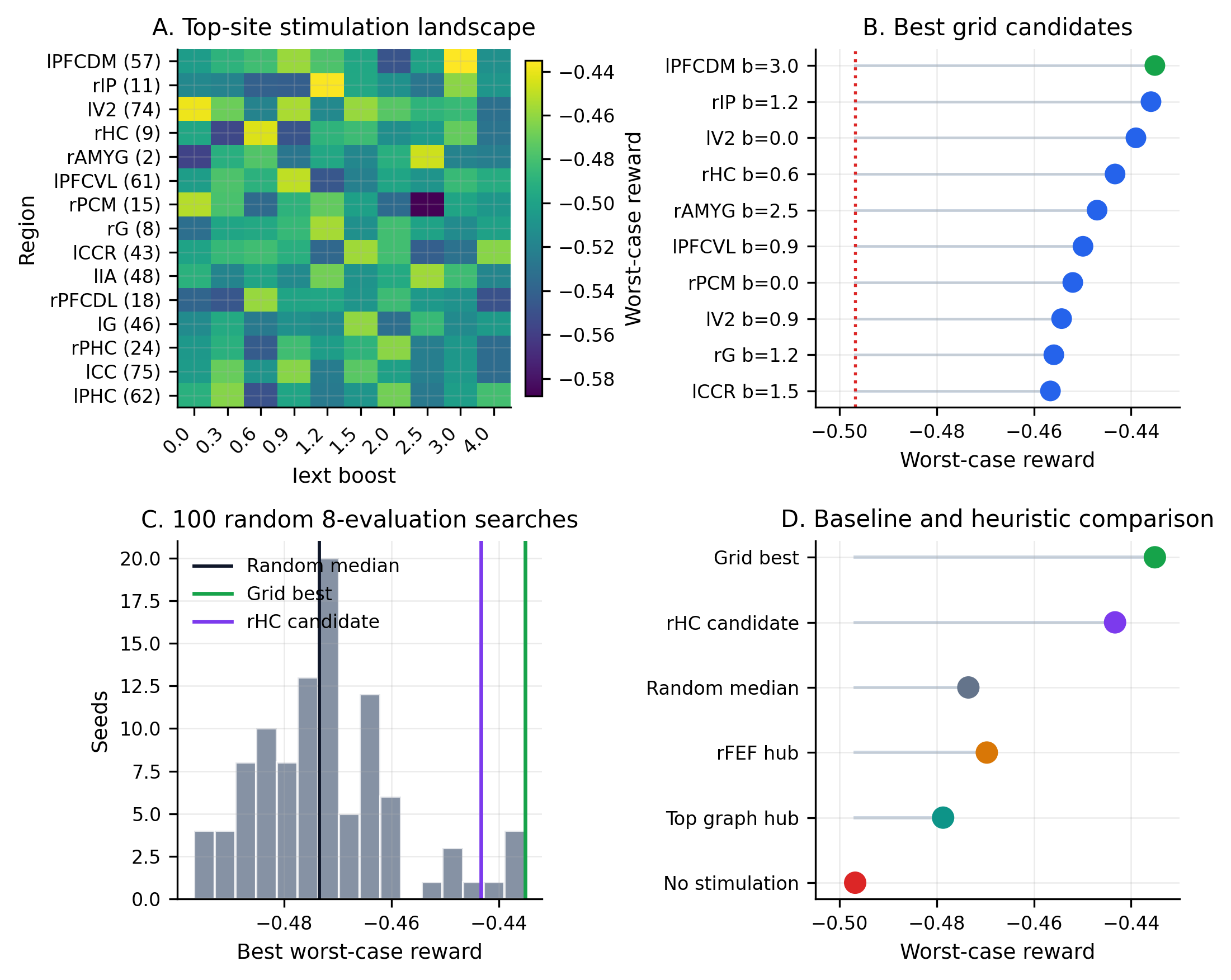}
  \caption{\textbf{External-stimulation landscape calibration.} \textbf{A}, Current-code grid search over the best 15 regions and 10 boost values. \textbf{B}, Top 10 grid candidates by worst-case reward. \textbf{C}, Distribution of best rewards from 100 random eight-evaluation searches compared with the grid optimum and rHC candidate. \textbf{D}, Current-code comparison of grid optimum, literature-selected rHC, random search, hub heuristics, and no stimulation.}
  \label{fig:landscape}
\end{figure}

The new landscape experiment provides the missing context for the constrained external-stimulation result (Figure~\ref{fig:landscape}). Across 760 evaluated region-boost candidates, the best grid point was $b=3.0$ at lPFCDM (index 57), with worst-case reward $-0.4351$ and mean reward $-0.4075$ (Table~\ref{tab:landscape}). The no-stimulation baseline in the same current-code environment had worst-case reward $-0.4968$, so the grid optimum improved the current-code minimax endpoint by 12.4\%.

The archived LLM-selected rHC protocol remains competitive in this calibrated landscape. At $b=0.6$, rHC ranked fourth of 760 candidates, with worst-case reward $-0.4433$, a 10.8\% improvement over the current-code no-stimulation baseline and only $0.0082$ reward units below the grid optimum. In contrast, the median best result among 100 random eight-evaluation searches was $-0.4735$, and graph-hub heuristics did not dominate: the best rFEF hub candidate reached $-0.4697$, while the top graph hub rPFCORB was best at no stimulation and reached only $-0.4787$. This does not prove that the LLM is a superior optimizer, but it substantially strengthens the narrow claim that a literature prior can land near high-performing regions in a named anatomical search space. The result is especially useful because the small grid-searchable space gives an independent audit of a proposal mechanism intended for larger spaces where a full grid would not be feasible.

\begin{table}[t]
  \centering
  \small
  \caption{\textbf{Current-code external-stimulation landscape.} All rows are evaluated in the same \texttt{uv} environment over the five-patient minimax cohort. Rewards are negative seizure-burden scores; less negative is better.}
  \label{tab:landscape}
  \begin{tabular}{llrrr}
    \toprule
    Method & Candidate & Evals. & Worst & Gap \\
    \midrule
    Exhaustive grid & lPFCDM, $b=3.0$ & 760 & $-0.4351$ & $0.0000$ \\
    LLM-selected rHC & rHC, $b=0.6$ & 8 & $-0.4433$ & $0.0082$ \\
    Random median & best of 8 & 8 & $-0.4735$ & $0.0384$ \\
    rFEF hub heuristic & rFEF, $b=2.5$ & 10 boosts & $-0.4697$ & $0.0346$ \\
    Top graph hub & rPFCORB, $b=0.0$ & 10 boosts & $-0.4787$ & $0.0436$ \\
    No stimulation & rHC, $b=0.0$ & 1 & $-0.4968$ & $0.0617$ \\
    \bottomrule
  \end{tabular}
\end{table}

\subsection{The 20-patient cohort exposed responder heterogeneity}
The 20-patient cohort is the most important translational check. Mean reward was $-0.4501$ at baseline and $-0.4518$ under the rHC protocol. The paired test was not significant ($t=-0.125$, $p=0.9019$), and the standardized effect size was negligible ($d=-0.049$). However, the aggregate null result hides strong patient-level heterogeneity: 11 of 20 virtual patients improved, while 9 worsened.

SOZ stratification suggests one possible explanation (Table~\ref{tab:soz}). The temporal subgroup had a positive mean reward change of $+0.0563$, while hippocampal, frontal, and occipital subgroups were near zero or negative. The temporal subgroup contains only two patients, so this should not be overinterpreted. Still, the pattern is biologically plausible enough to motivate follow-up: stimulation at a hippocampal/temporal target may help a subset of temporal-network patients while being neutral or adverse elsewhere.

\begin{table}[t]
  \centering
  \caption{\textbf{SOZ-stratified virtual cohort response to rHC stimulation.} Values are optimized minus baseline reward. Positive values indicate lower simulated seizure burden under stimulation.}
  \label{tab:soz}
  \begin{tabular}{lrrr}
    \toprule
    SOZ class & Patients & Mean reward change & SEM \\
    \midrule
    Temporal & 2 & $+0.0563$ & $0.0297$ \\
    Hippocampal & 9 & $-0.0078$ & $0.0191$ \\
    Frontal & 4 & $-0.0043$ & $0.0302$ \\
    Occipital & 5 & $-0.0119$ & $0.0363$ \\
    \bottomrule
  \end{tabular}
\end{table}

\subsection{Network topology cautions against simple hub-targeting}
\begin{figure}[t]
  \centering
  \includegraphics[width=\linewidth]{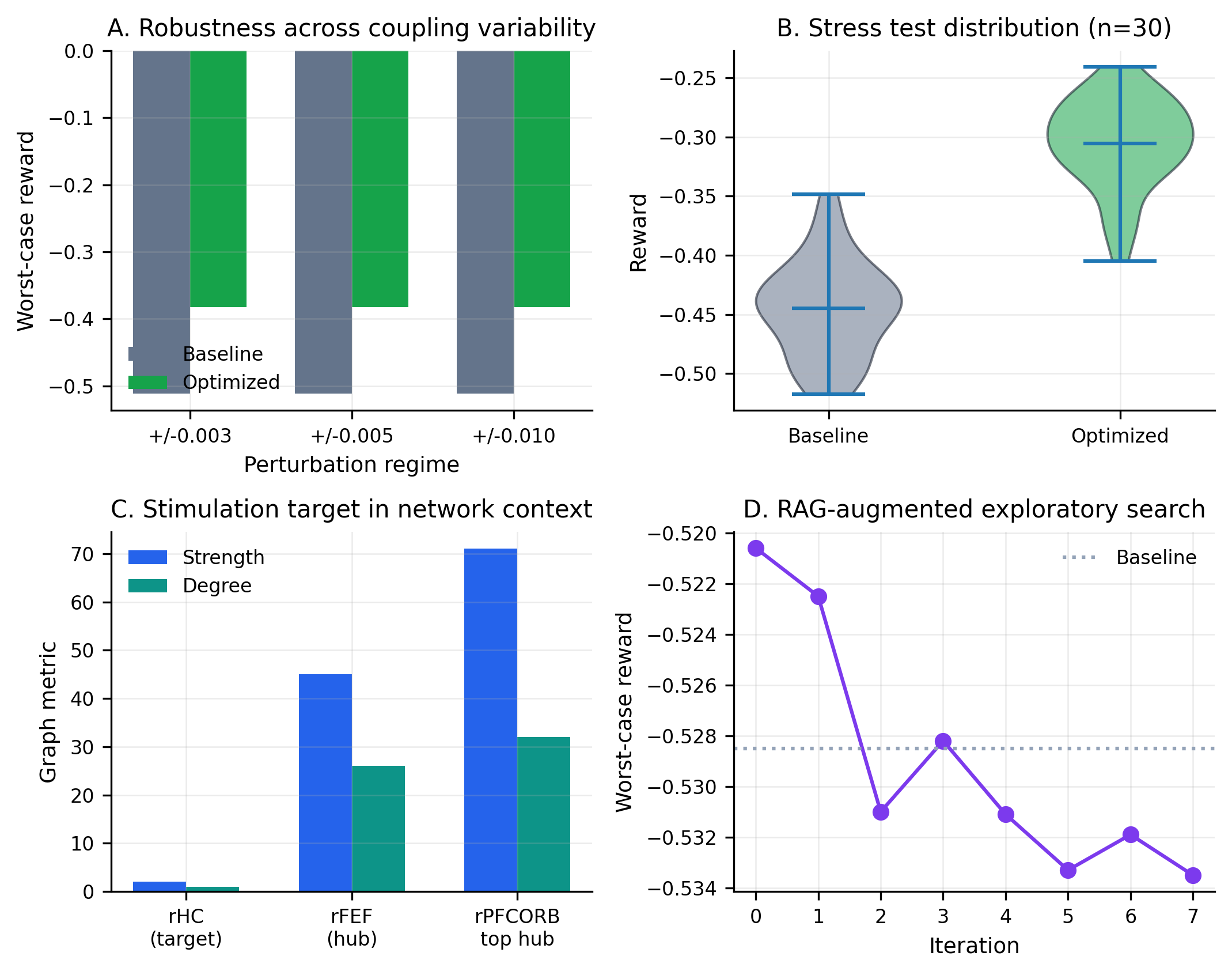}
  \caption{\textbf{Robustness, topology, and retrieval-augmented search.} \textbf{A}, The optimized intrinsic regime preserves worst-case improvement across archived perturbation regimes. \textbf{B}, Stress-test reward distributions remain shifted toward improved reward under optimized intrinsic parameters. \textbf{C}, The selected rHC stimulation target is far less connected than prefrontal hubs, cautioning against simple graph-hub heuristics. \textbf{D}, RAG-augmented search provided rationales and citations but did not exceed the best simpler LLM-guided external-stimulation result.}
  \label{fig:robustness}
\end{figure}

The topology artifact adds a useful negative result. The best clinical-style target, rHC, is not a network hub in the loaded connectome: its degree is 1 and strength is 2.0. By contrast, prefrontal hubs such as rPFCORB have degree 32 and strength 71.0. A pure hub-targeting heuristic would therefore have avoided rHC. Yet the archived LLM and RAG trajectories repeatedly returned to hippocampal stimulation because of the epilepsy literature prior and observed rewards (Figure~\ref{fig:robustness}). This tension is important. Effective targets may be mechanistically local, topologically central, or state-dependent; whole-brain stimulation design should test all three hypotheses rather than assume that graph hubs are automatically optimal.

The RAG-augmented search did not outperform the simpler LLM-guided clinical search in the archived results. Its best reward was $-0.5206$, a 1.5\% gain over baseline, and the best proposal appeared at the first iteration. Retrieval added interpretability by surfacing papers and mechanistic rationales, but it did not by itself solve the constrained optimization problem.

\section{Discussion}
This study demonstrates that literature-guided minimax search can identify robust dynamical regimes in virtual epilepsy models. The intrinsic-control result is the clearest evidence: across the archived optimization trace, held-out evaluation, and stress distribution, the selected parameter regime consistently reduces the simulated seizure-burden score. Because the objective is worst-case reward, the result is stronger than an average-case improvement: it explicitly improves the least favorable sampled virtual patient.

The translational story is more constrained. External stimulation at a single region produced only modest gains, and the 20-patient cohort did not show an aggregate benefit. This should not be read as a failure of the pipeline. It is precisely the kind of stress test a robust neurostimulation workflow should expose. The method found that direct model-control parameters can produce a large effect, then showed that the currently tested clinical proxy captures only a small portion of that effect and may worsen some virtual patients. In a clinical design setting, that information is valuable because it redirects attention toward patient selection, target stratification, waveform design, and closed-loop timing.

The landscape experiment sharpens the LLM interpretation. If the search space can be exhaustively enumerated, a grid is the better scientific instrument. The LLM is interesting because most realistic design spaces are not like that. A practical stimulation optimizer must reason over anatomical targets, device constraints, waveform families, closed-loop timing, patient states, and literature-derived mechanistic priors. Many of those variables are discrete, conditional, or semantic, and the cost of simulating every combination grows quickly. The 760-candidate landscape is therefore a validation sandbox: it shows that the literature-guided proposal was not arbitrary, while preserving an external reference point that would disappear in a richer design space.

The results also clarify the role of LLMs in computational neuroscience. The LLM is useful as a proposal engine and literature interface, not as a source of truth. It can rapidly suggest plausible targets such as hippocampal stimulation for temporal-lobe epilepsy, explain why a candidate is mechanistically reasonable, and adapt proposals based on previous rewards. But every scientific claim in this workflow still depends on simulation output, artifact inspection, and domain validation. The architecture is therefore best understood as a constrained human-readable hypothesis generator wrapped around a conventional computational model.

\subsection{Limitations}
Several limitations are substantial. First, the virtual cohort is small, and the patient variability model is simple. Coupling noise and synthetic SOZ categories are not substitutes for patient-specific diffusion MRI, stereo-EEG, seizure diaries, medication history, or implanted-device recordings. Second, the reward function is a proxy: negative variance of the simulated fast Epileptor state is useful for computational comparison but is not a validated endpoint for seizure frequency, severity, cognition, or quality of life. Third, the intrinsic-control result is not directly clinically actionable because changing $x_0$ and global coupling is a model intervention, not a neurosurgical protocol. Fourth, the literature-mining pipeline uses LLM-generated extraction, clustering, and scoring without expert adjudication. Its outputs should be treated as structured hypotheses. Fifth, the external-stimulation model abstracts away pulse width, frequency, timing, electrode geometry, tissue conductivity, stimulation artifacts, and safety constraints that dominate real devices.

Additional implementation limitations follow from the current repository artifacts. The simulation code uses a deterministic integrator and does not model observation noise, neural state uncertainty, or stochastic seizure triggering. The external-current boost is a coarse input perturbation, not a biophysical model of implanted stimulation. The eight-evaluation optimization budget and single archived LLM trace are too small to establish optimizer superiority; repeated LLM runs with fixed prompts, model settings, and seeds would be needed for that claim. The all-region landscape reduces one major uncertainty by showing where the selected rHC candidate sits in the bounded search space, but it does not validate the candidate in richer waveform, timing, or patient-specific spaces. Finally, the held-out and stress-test artifacts increase confidence that the intrinsic-control result is not confined to the original five samples, but the repository does not record enough run provenance to treat them as formally pre-specified validation cohorts.

\subsection{Future work}
The next version of this project should move from sampled virtual cohorts to patient-specific digital twins, with structural connectomes, seizure onset hypotheses, and validation against held-out electrophysiology. The stimulation model should include waveform, frequency, pulse width, phase, and closed-loop timing, because the external-current boost used here is too coarse to represent modern responsive neurostimulation. Those additions would make exhaustive grid search impractical, which is exactly where proposal engines become scientifically useful. The optimization layer should therefore compare repeated LLM-guided proposal runs against stronger baselines, including random search, evolutionary strategies, constrained Bayesian optimization, expert-designed protocols, and hybrid approaches that use grids only for small subspaces. Finally, the literature-mining stage should be reviewed by epilepsy clinicians and computational neuroscientists before it is used to prioritize translational experiments.

\section{Conclusion}
A robust neurostimulation optimizer should optimize for the patient who fails first, not the average patient who improves most easily. This project operationalizes that principle by coupling literature-guided LLM proposals with TVB Epileptor minimax evaluation. The intrinsic model-control experiment produced a large positive-control improvement in worst-case reward, while the external-stimulation experiments exposed the expected difficulty of translating that control signal into a protocol-like intervention. The all-region landscape makes the LLM result more interpretable: the selected rHC candidate was not the grid optimum, but it landed near the top of a 760-candidate anatomical search space with few evaluations. The main contribution is therefore not a claim of clinical efficacy, but a disciplined framework for using small auditable simulation spaces to validate proposal mechanisms intended for larger, non-exhaustive neurostimulation design problems.

\section*{Code and Data Availability}
All code, static result artifacts, figures, and the public GitHub Pages essay are available at \url{https://github.com/liuzhitong330/tvb-llm-robust-neurostim}. The figures in this manuscript are generated from checked-in JSON artifacts by \texttt{paper/generate\_figures.py}. The paper does not introduce new simulation results beyond those artifacts.

\section*{Ethics and Clinical Use Statement}
This work uses simulated virtual patients and contains no human-subject data. It is not medical advice, clinical guidance, or evidence that any stimulation protocol is safe or effective in patients.

\section*{Acknowledgments}
This manuscript describes Cathy Liu's public research project on LLM-guided robust optimization for epilepsy neurostimulation using The Virtual Brain.

\bibliographystyle{unsrt}

\begin{thebibliography}{9}

\bibitem{sanzleon2013tvb}
P. Sanz Leon, S. A. Knock, M. M. Woodman, L. Domide, J. Mersmann, A. R. McIntosh, and V. Jirsa.
\newblock The Virtual Brain: a simulator of primate brain network dynamics.
\newblock \emph{Frontiers in Neuroinformatics}, 7:10, 2013.

\bibitem{jirsa2014seizure}
V. K. Jirsa, W. C. Stacey, P. P. Quilichini, A. I. Ivanov, and C. Bernard.
\newblock On the nature of seizure dynamics.
\newblock \emph{Brain}, 137(8):2210--2230, 2014.

\bibitem{proix2017vep}
T. Proix, V. K. Jirsa, F. Bartolomei, M. Guye, and W. Truccolo.
\newblock The Virtual Epileptic Patient: individualized whole-brain models of epilepsy spread.
\newblock \emph{NeuroImage}, 145:377--388, 2017.

\bibitem{morrell2011rns}
M. J. Morrell and the RNS System in Epilepsy Study Group.
\newblock Responsive cortical stimulation for the treatment of medically intractable partial epilepsy.
\newblock \emph{Neurology}, 77(13):1295--1304, 2011.

\bibitem{fisher2010sante}
R. Fisher, V. Salanova, T. Witt, R. Worth, T. Henry, R. Gross, K. Oommen, I. Osorio, J. Nazzaro, D. Labar, and others.
\newblock Electrical stimulation of the anterior nucleus of thalamus for treatment of refractory epilepsy.
\newblock \emph{Epilepsia}, 51(5):899--908, 2010.

\bibitem{baud2018rhythms}
M. O. Baud, J. K. Kleen, E. A. Mirro, J. C. Andrechak, D. King-Stephens, E. F. Chang, and V. R. Rao.
\newblock Multi-day rhythms modulate seizure risk in epilepsy.
\newblock \emph{Nature Communications}, 9:88, 2018.

\bibitem{snoek2012bayesopt}
J. Snoek, H. Larochelle, and R. P. Adams.
\newblock Practical Bayesian optimization of machine learning algorithms.
\newblock In \emph{Advances in Neural Information Processing Systems}, 2012.

\end{thebibliography}

\end{document}